# Title: Alignment-Aware and Reliability-Gated Multimodal Fusion for Unmanned Aerial Vehicle Detection Across Heterogeneous Thermal–Visual Sensors

**Authors:** Ishrat Jahan[1], Molla E Majid[2], M Murugappan[3,4*], Muhammad E. H. Chowdhury[5*], N.B.Prakash[6], Saad Bin Abul Kashem[7], Balamurugan Balusamy[8*], Amith Khandakar[5]

**Affiliation:**
[1] Department of Computer Science and Engineering, Shahjalal University of Science & Technology, Sylhet, Bangladesh
[2] Computer Applications Department, Academic Bridge Program, Qatar Foundation, Doha, Qatar.
[3] Intelligent Signal Processing (ISP) Research Lab, Department of Electronics and Communication Engineering, Kuwait College of Science and Technology, Block 4, Doha, Kuwait
[4] Department of Electronics and Communication Engineering, Vels Institute of Sciences, Technology, and Advanced Studies, Chennai, Tamilnadu, India.
[5] Department of Electrical Engineering, Qatar University, Doha 2713, Qatar.
[6] School of Computing Science and Engineering, Vellore Institute of Technology Bhopal University, Bhopal, India.
[7] Department of Computing Science, AFG College with the University of Aberdeen, Doha, Qatar
[8] School of Engineering and IT, Manipal Academy of Higher Education, Dubai Campus, Dubai, UAE

Corresponding Author (s):
**Prof. Dr. Balamurugan Balusamy**
School of Engineering and IT,
Manipal Academy of Higher Education,
Dubai Campus, Dubai,
UAE
Email: balamurugan.balusamy@dxb.manipal.edu

**M Murugappan**
Intelligent Signal Processing (ISP) Research Lab,
Department of Electronics and Communication Engineering,
Kuwait College of Science and Technology,
Block 4, Doha,
Kuwait
Email: m.murugappan@kcst.edu.kw

Muhammad E.H.Chowdhury
Department of Electrical Engineering,
Qatar University,
Doha 2713,
Qatar.
Email: mchowdhury@qu.edu.qa

# Alignment-Aware and Reliability-Gated Multimodal Fusion for Unmanned Aerial Vehicle Detection Across Heterogeneous Thermal–Visual Sensors

**Abstract**

Reliable unmanned aerial vehicle (UAV) detection is critical for autonomous airspace monitoring but remains challenging when integrating sensor streams that differ substantially in resolution, perspective, and field of view. Conventional fusion methods—such as wavelet-, Laplacian-, and decision-level approaches—often fail to preserve spatial correspondence across modalities and suffer from annotation of inconsistencies, limiting their robustness in real-world settings. This study introduces two fusion strategies, Registration-aware Guided Image Fusion (RGIF) and Reliability-Gated Modality-Attention Fusion (RGMAF), designed to overcome these limitations. RGIF employs Enhanced Correlation Coefficient (ECC)-based affine registration combined with guided filtering to maintain thermal saliency while enhancing structural detail. RGMAF integrates affine and optical-flow registration with a reliability-weighted attention mechanism that adaptively balances thermal contrast and visual sharpness. Experiments were conducted on the Multi-Sensor and Multi-View Fixed-Wing (MMFW)-UAV dataset comprising 147,417 annotated air-to-air frames collected from infrared, wide-angle, and zoom sensors. Among single-modality detectors, YOLOv10x demonstrated the most stable cross-domain performance and was selected as the detection backbone for evaluating fused imagery. RGIF improved the visual baseline by 2.13% mAP@50 (achieving 97.65%), while RGMAF attained the highest recall of 98.64%. These findings show that registration-aware and reliability-adaptive fusion provides a robust framework for integrating heterogeneous modalities, substantially enhancing UAV detection performance in multimodal environments.

**1. Introduction**

Airspace is the regulated region of the atmosphere used for aviation activities, which includes both manned and unmanned operations. UAVs (unmanned aerial vehicles) are becoming more common because they are cheap, easy to use, and can be used in a variety of civil and commercial settings [1]. Their ability to take clear pictures and track things accurately has led to widespread application in areas like farming, air quality testing, and watching the environment [2]. The quick rise in the use of UAVs, on the other hand, caused serious security concerns, particularly about the possibility of collisions, entering airspace without permission, and keeping people safe [3-5]. Recent events have shown how dangerous illegal UAV activity may be for operations and security. The 2018 murder attempt on Venezuelan President Nicolas Maduro using explosive drones and the Gatwick Airport closure that affected over 140,000 people both show how UAVs can be misused to put public safety and important infrastructure at risk [6, 7]. Public trust depends on strict safety rules in regulated airspace. Without them, UAV integration will fail. We also face serious privacy concerns because these devices can record sensitive data. That capability forces us to build stronger regulatory and technical defenses [8-10]. Regulatory measures such as airspace classification, altitude restrictions, operator licensing, and geofencing technologies have improved UAV management and airspace safety [11, 12]. However, these frameworks remain largely rule-based and reactive, relying on static constraints that cannot adapt to changing flight conditions [13]. Static protocols prove inadequate as UAV operations become denser and more diverse. These rigid frameworks struggle with real-time hazard detection and fail to coordinate effectively with manned flight [14, 15]. Current limitations necessitate intelligent, data-driven systems. These architectures must combine multimodal perception, advanced sensing, and autonomous decision-making to maintain safety and support large-scale UAV integration.

Limitations in rule-based safety systems have shifted research toward data-driven perception frameworks. These systems function autonomously within dynamic airspace. Computer vision offers a scalable and cost-effective solution. It enables UAV detection and tracking through real-time image analysis instead of static operational rules [16, 17]. Recent advances in deep learning have transformed computer vision by enabling models to learn complex spatial and spectral representations directly from data [18-20]. Deep architectures now serve as the core of most vision-based UAV perception pipelines [21, 22]. Object detection plays a fundamental role in this context by providing situational awareness for navigation, collision avoidance, and target identification [23, 24].

However, prior research largely favors model architecture over input quality or multimodal sensing. In UAV detection, physical factors such as low signal-to-noise ratios in daylight, occlusion, and poor visibility under low light make single-sensor data unreliable. Thermal (infrared, IR) imagery provides stable contrast signatures that are largely independent of illumination conditions, while visual (RGB) imagery captures fine structural details. Fusing these complementary modalities enhances detection robustness across varying environmental conditions. Existing studies have explored either sensor-level fusion, which combines images before model training, or feature-level fusion, which merges learned representations within neural networks. For example, Cheng et al. [25] proposed SLBAF-Net to address UAV detection in difficult weather conditions using feature-level fusion. While this framework confirms the value of multimodal integration, the study relied on a limited dataset. Most existing methods still assume homogeneous image resolutions and overlook alignment issues between heterogeneous sensors. These limitations necessitate more adaptive fusion mechanisms for UAV perception.

Aydin et al. [26] employed YOLOv5 with pre-trained weights on a dataset of 2,395 images, including 1,916 drone and 479 bird images collected from public sources. Their model achieved a mAP of 90.40%. However, the dataset size was small and lacked diversity, making the results less reliable. Seidaliyeva et al. [27] reviewed UAV detection methods, including radar, RF, acoustic, vision-based, and sensor fusion approaches. While the survey was comprehensive, traditional techniques such as RF and acoustic are now widely regarded as less effective and outdated compared to modern deep learning–based vision methods. Zamri et al. [28] employed a dataset of 2,970 images to differentiate drones from drone-like objects such as birds and proposed P2-YOLOv8n-ResCBAM, which integrated attention modules and an additional detection head. Their model achieved a mAP of 92.6%. However, the dataset size was relatively small, and the performance gain came at the cost of increased model complexity, limiting scalability compared to larger object detection benchmarks. Hakani et al. [6] assembled a dataset of 9,995 images from online sources and their own drone flights, and applied YOLOv9 with transfer learning on the NVIDIA Jetson Nano for real-time drone detection. Their model achieved a mAP of 95.7%. However, the dataset was heterogeneous and partly reused from general-purpose sources like COCO, the reliance on Jetson Nano limits scalability. Wisniewski et al. [29] developed a Faster R-CNN model trained solely on a synthetic dataset and evaluated it on the MAV-Vid real-world drone dataset, achieving an AP50 of 97.0%. However, the reliance on synthetic data raises concerns about sim-to-real transferability, and the narrow evaluation of a single dataset limits the generalizability of their findings to broader drone detection scenarios. There are relatively few studies on multimodal fusion and works specifically addressing UAV detection remain extremely scarce. Wei et al. [30] introduced a YOLOv8-based heterogeneous infrared–visible fusion model (YOLOv8-CVIFB) for power equipment inspection. Their framework effectively integrated registration, fusion, and detection. However, the dataset was limited, and the study did not address UAV detection. Similarly, Zhoufeng Yu et al. [31] proposed an optical–SAR image fusion approach using SIFT feature extraction and entropy-based evaluation. This method achieved recognition accuracy, but it did not target UAV detection.

A major limitation of existing studies on UAV detection is their reliance on limited datasets, synthetic data, or an emphasis on model-level enhancements. Architectural modifications alone yield only marginal improvements in the advancing era of object detection and YOLO models. Substantial progress can instead be achieved through diverse datasets and thorough preprocessing strategies. Incorporating multi-temporal, multi-sensor, and multi-view imagery provides a more comprehensive representation of drones. This study bridges the gap in prior research by utilizing an openly accessible multi-sensor, multi-view dataset of fixed-wing UAVs curated for air-to-air vision tasks. We introduce a novel cross-sensor fusion framework that integrates heterogeneous thermal and visual cues before detection. This approach enables structurally enhanced representations that substantially strengthen downstream UAV recognition. YOLO-based detectors have consistently demonstrated state-of-the-art performance across a wide range of object detection tasks. In this work, YOLOv10x is employed as the detection backbone due to its favorable accuracy–efficiency balance and suitability for real-time UAV recognition. Thermal (infrared) and visual (wide view) modalities were first used independently to train and assess the baseline detection models. To improve generalization across viewing conditions, the visual detector was further fine-tuned using the zoom-view samples available in the dataset. However, cross-modal fusion is non-trivial due to the heterogeneity of infrared and visual sensors, which exhibit substantial differences in spatial resolution, radiometric behavior, and field of view.

In practical UAV-mounted sensing systems, such heterogeneity results in RGB and infrared inputs that are not only radiometrically distinct but also spatially mismatched in resolution and alignment. Most existing fusion approaches implicitly assume homogeneous or uniformly resized inputs, which limit their applicability to real-world UAV deployments. This motivates fusion strategies that explicitly accommodate heterogeneous sensor characteristics rather than relying on pre-alignment or uniform resizing. To mitigate these inconsistencies, several pixel-level and feature-level fusion strategies were examined to ensure coherent multimodal integration.

Despite significant progress current UAV detection models often struggle to align data across modalities with different spatial characteristics due to assumptions of homogeneous image resolutions. To overcome this limitation, our work introduces fusion strategies designed to handle heterogeneous resolutions (1024×1280 infrared vs. 2160×3840 visual) while maintaining geometric and semantic consistency. We propose two multimodal integration strategies: Registration-aware Guided Image Fusion (RGIF) and Reliability-Gated Modality-Attention Fusion (RGMAF). These mechanisms address cross-sensor misalignment and modality reliability to create a robust framework capable of integrating heterogeneous cues for enhanced UAV detection. The key contributions of this study are as follows:

- Registration-aware Guided Image Fusion (RGIF): A fusion strategy that preserves complementary thermal and visual information under heterogeneous sensor resolutions and imperfect cross-modal alignment through ECC-based alignment and guided filtering.
- Reliability-Gated Modality-Attention Fusion (RGMAF): An adaptive fusion mechanism that regulates modality contributions using reliability-weighted attention to improve robustness under varying illumination and sensing conditions.
- A unified UAV detection framework: A comprehensive multimodal detection pipeline evaluated across thermal, visual, and fused modalities, establishing a benchmark for UAV perception under heterogeneous sensor conditions.

This paper is organized as follows. **Section 2** reviews traditional approaches to image fusion. **Section 3** describes the dataset and the proposed fusion methodologies. **Section 4** presents experimental results. **Section 5** discusses the findings, limitations, and directions for future work. **Section 6** concludes the paper, and **Section 7** lists all references.

## 2. Background

Traditionally, infrared and visible image fusion can be grouped into four major categories: pixel-level, multi-scale transform, filter-based, and color-space-based techniques. Pixel-level fusion directly combines pixel intensities from both modalities, with common examples including alpha blending, weighted averaging, and overlay fusion [32]. Multi-scale transform–based fusion decomposes images into low- and high-frequency components (e.g., Laplacian pyramid or wavelet transform), which are fused and then reconstructed to preserve both saliency and structural details. Earlier studies have also extended this framework by incorporating sparse representation to enhance detail preservation [33]. Filter-based fusion employs edge-preserving filters, such as the guided filter, to separate base and detail layers that are then selectively integrated [34]. Finally, color space-based fusion enhances luminance with thermal information while preserving visual chroma, for instance by operating in the YCrCb color space.

### 2.1. Pixel-Level Fusion Method

#### A. Alpha Blending

Alpha blending fuses the thermal and visual modalities through a simple convex linear combination $F=\alpha I+(1-\alpha)V$. Although computationally efficient, this method often suppresses salient target features because it applies fixed weights across the entire image without considering local structures [35]. It performs poorly when fusing heterogeneous resolutions since it assumes identical pixel correspondence between modalities.

#### B. Weighted Averaging

Weighted averaging extends alpha blending by assigning modality-specific weights that emphasize either thermal saliency or visual detail depending on the task. While it provides slightly better balance between modalities, it still lacks adaptivity and struggles to capture fine structural details [36].

#### C. Overlay Fusion

Overlay fusion maps the thermal modality directly onto the visual image, often using pseudo-color to highlight target regions within the visual context. This approach improves interpretability for human observation but introduces distortions and is less suitable for quantitative detection tasks [32].

### 2.2. Multi-Scale Transform–Based Fusion Methods
#### A. Laplacian Pyramid
In Laplacian pyramid fusion, both modalities are decomposed into multi-scale layers. Low-frequency components are typically averaged, while high-frequency components are combined using selection rules (e.g., maximum gradient) [37]. While it preserves both saliency and edge detail, reconstruction becomes unreliable when resolution disparity causes inconsistent gradient magnitudes.

#### B. Wavelet Transform
Wavelet-based fusion uses the discrete wavelet transform (DWT) to combine approximation and detail coefficients [38]. It retains texture and contrast but exhibits aliasing and coefficient mismatch when input resolutions or aspect ratios differ.

### 2.3. Filter-Based Fusion Methods
#### A. Guided Filter
Guided filter fusion relies on an edge-preserving filter to separate images into base and detail layers. Thermal information is emphasized in the base layer, while structural edges from the visual modality are transferred through the guidance map [39]. It produces sharp fused outputs on aligned data but is highly sensitive to mis-registration and local motion between modalities.

### 2.4. Color-Space–Based Fusion Methods
#### A. YCrCb Luminance
In YCrCb luminance fusion, the visual image is converted into the YCrCb color space, and the thermal modality replaces or is integrated into the luminance channel. The chroma channels (Cr, Cb) from the visual image are then recombined to form the final fused image [34]. This enhances brightness perception but suffers severe color bleeding when sensor resolutions or viewpoints are inconsistent. **Table 1** presents a summary of classical fusion techniques.

Table 1: Summary of Classical Fusion Techniques

| Category | Typical Methods | Computational Complexity | Alignment Sensitivity | Limitation under Heterogeneous Inputs |
|---|---|---|---|---|
| **Pixel-level** | Alpha blending, weighted averaging, overlay | Low | High | Requires identical pixel grids; misalignment causes ghosting. |
| **Multi-scale transform** | Laplacian, Wavelet | Moderate–High | Moderate | Coefficient mismatch under different resolutions. |
| **Filter-based** | Guided filter | Moderate | High | Dependent on accurate local correspondence. |
| **Color-space-based** | YCrCb luminance | Low–Moderate | Very High | Distorted color mapping with mis-registered modalities. |

### 2.5 Modern Deep Learning–Based Fusion Baselines
Recent studies have also explored deep learning–based fusion frameworks such as DenseFuse [40] and FusionGAN [41], which learn end-to-end fusion mappings using encoder–decoder and adversarial architectures, respectively. These models achieve visually appealing results on aligned datasets but often degrade when applied to heterogeneous infrared–visible inputs. Their reliance on co-registered, same-resolution images further highlights the need for registration-aware and reliability-adaptive fusion approaches, as addressed in this study.

## 3. Methodology
### 3.1. Dataset Description

The dataset used in this study is the publicly available MMFW-UAV dataset [42], which contains multi-sensor and multi-view imagery of fixed-wing unmanned aerial vehicles (UAVs), intended for air-to-air object detection research. It was collected using three onboard sensors: a 48-megapixel RGB zoom camera, a 12-megapixel RGB wide-angle camera, and an uncooled vanadium oxide (VOx) thermal imaging sensor. The full dataset comprises 147,417 images, with an equal number of images (49,139) from each sensor type. Data acquisition was conducted over twelve UAV sorties using platforms of varying sizes. The UAVs differ in physical dimensions and flight speeds, with some variants also equipped with solar panel casings, introducing appearance variability across modalities. Flights were conducted at altitudes ranging from 25 m to 100 m, and images were captured from three viewpoints: top-down, horizontal, and bottom-up, under diverse illumination conditions including morning, afternoon, and cloudy weather. The wide camera had an 84° diagonal field of view, the zoom camera operated within a narrower 21–75 mm equivalent range, and the thermal camera had a 61° diagonal field of view. Each image contains a single fixed-wing UAV instance, annotated with axis-aligned bounding boxes in both Pascal VOC (xmin, ymin, xmax, ymax) and MS COCO (xc, yc, w, h) formats using one category label ("Fixed_Wing_UAV"). This structure supports the development and evaluation of object detection models under multimodal and multi-view conditions. **Figure 1** presents sample images from the dataset and **Table 2** summarizes the per-modality statistics.

Table 2: Per-modality sensor specifications and image characteristics of the MMFW-UAV dataset

| Sensor Type | Resolution (px) | Bit Depth | Field of View | Lens / Sensor | #Images | Aspect Ratio | Typical Illumination | Notes |
|---|---|---|---|---|---|---|---|---|
| RGB Zoom | 3840 × 2160 | 8-bit | Not reported (only focal length given) | 21–75 mm zoom lens, 1/2″ CMOS | 49,139 | 16:9 | Daytime | High-detail visible-light imagery |
| RGB Wide-angle | 3840 × 2160 | 8-bit | 84° diagonal | 4.5 mm prime lens | 49,139 | 16:9 | Daytime | Large FoV, more environmental context |
| Thermal (VOx) | 1280 × 1024 | 8-bit | 61° diagonal | 9.1 mm prime lens | 49,139 | 5:4 | Low-light / cloudy | Grayscale thermal signatures |
| **Fused Image (RGIF / RGMAF)** | **1280 × 1024** | **8-bit (RGB mapped to thermal domain)** | **61° diagonal (thermal FoV)** | **Visual warped into thermal geometry** | **49,139** | **5:4** | **Effective across day/night** | **RGB warped to thermal frame, fusion preserves thermal structure** |

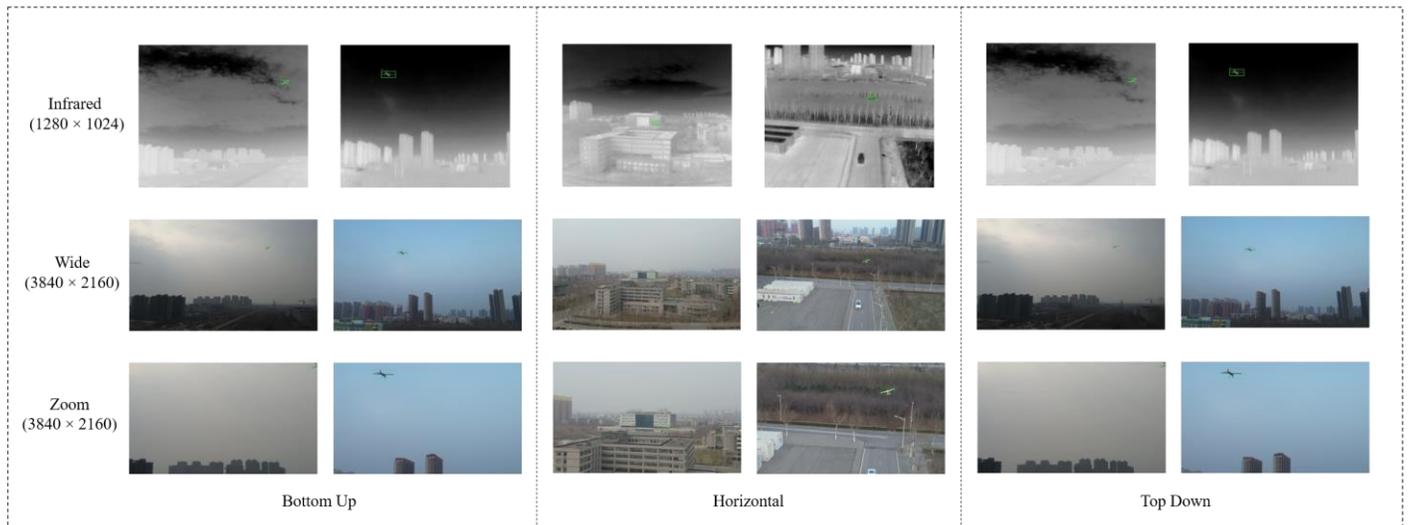

Figure 1: Representative sample images from the dataset

## 3.2. Proposed Method

In this study, we introduce two fusion strategies specifically designed to operate on thermal and visual images with mismatched spatial resolutions. Specifically, the thermal modality was represented as tensors of shape ($1024 \times 1280 \times 1$), while the visual modality was represented as tensors of shape ($2160 \times 3840 \times 3$). Within the visual domain, wide-view images were selected as the primary modality because zoom-view samples occasionally cropped objects of interest and were therefore unsuitable as independent training data. Two multimodal fusion modules were developed: Registration-aware Guided Image Fusion (RGIF), which performs pixel-level alignment through guided filtering, and Reliability-Gated Modality-Attention Fusion (RGMAF), which adaptively weights modality contributions based on estimated reliability. Both modules were constructed to integrate complementary thermal–visual cues while maintaining structural consistency across modalities. Handling multi-resolution imagery constituted a core challenge, as achieving cross-sensor correspondence required explicit registration and careful design of the feature alignment process.

Our experimental framework incorporated both multimodal fusion pipelines and single-sensor baselines. For the unimodal setting, several YOLO variants were trained and evaluated independently across thermal and visual modalities. YOLOv10x was selected as the detection backbone because it offered the most favorable accuracy–efficiency trade-off among the evaluated architectures, exhibiting higher representational capacity and lower computational cost compared with YOLOv9e and YOLOv12x. This balance renders YOLOv10x well-suited for real-time multimodal UAV detection. In addition, cross-sensor fine-tuning was performed by refining the wide-view visual detector using zoom-view samples to improve robustness to UAV appearance variations across viewing conditions. Within the fusion stage, thermal and visual feature maps were explicitly aligned to preserve spatial correspondence prior to integration.

The overall pipeline consisted of generating aligned multimodal inputs via image-level fusion, followed by five-fold cross-validation using disjoint training, validation, and test splits. Detection performance was benchmarked across multiple state-of-the-art YOLO architectures to provide consistent comparisons. **Figure 2** illustrates the overall workflow of the proposed methodology.

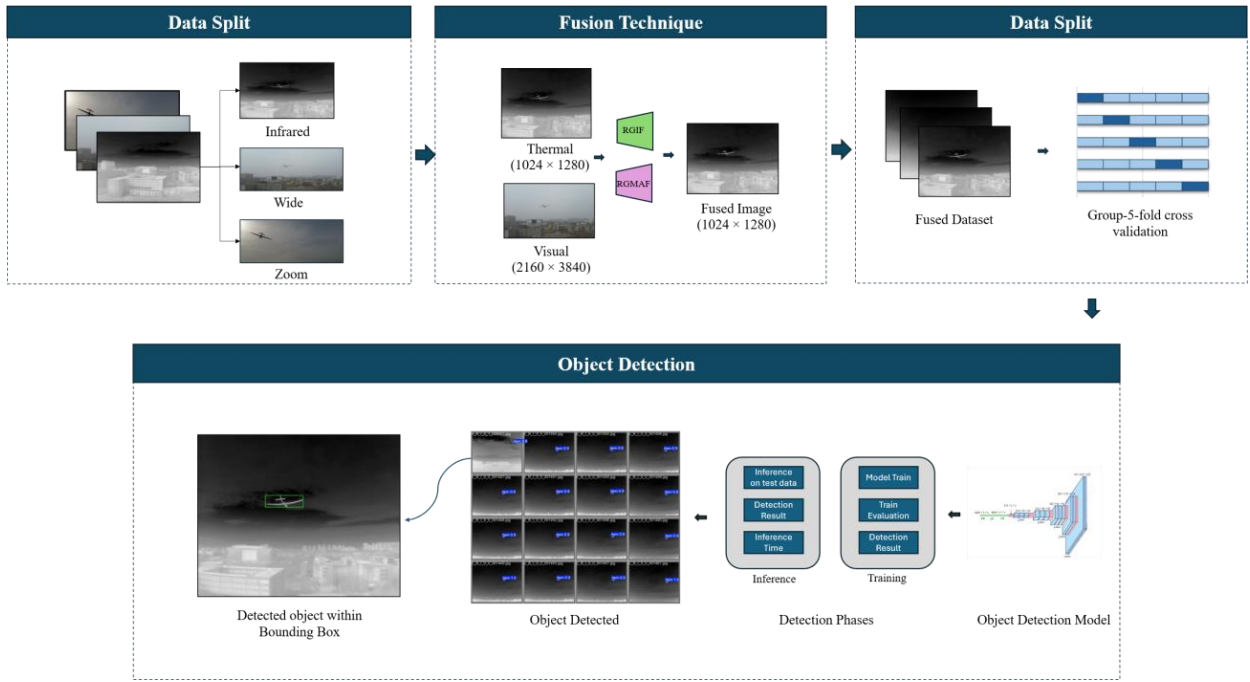

Figure 2: Workflow of the proposed methodology

## 3.3. Proposed Fusion Techniques
### 3.3.1. Registration-aware Guided Image Fusion (RGIF)

The fusion of thermal and visual modalities was implemented through a two-stage process comprising geometric registration followed by guided filtering. The thermal images (1024×1280×1) and wide-view visual images (2160×3840×3) were both upsampled by a factor of two to reduce interpolation artifacts and improve feature correspondence. The visual frames were then resized to match the thermal grid before registration. Each thermal image was min–max normalized to 8-bit intensity, ensuring uniform dynamic range across scenes. For spatial alignment, we employed the Enhanced Correlation Coefficient (ECC) method implemented in OpenCV to estimate an affine warp $W$ that maximizes the intensity correlation between the thermal reference T and the visual frame V:

$$W^* = arg_w \max \text{corr}(T(x,y), V(W(x,y))) \tag{1}$$

The ECC optimization was run for 100 iterations with a termination criterion of $10^{-5}$. If convergence failed, the visual frame was passed through without warping to maintain dataset completeness.

Following registration, the fused image was generated using a guided filtering strategy in which the visual grayscale image served as the guidance signal $I$, and the thermal image acted as the filtering input $p$. In each local window $\Omega_k$ of size r × r (with r=8), the model

$$q_i = a_k I_i + b_k, \quad \forall i \in \Omega_k \tag{2}$$

was applied, where

$$a_k = \frac{\text{cov}(I,p)}{\text{var}(I)+\varepsilon}, \quad b_k = \bar{P}_k - a_k \bar{I}_k \tag{3}$$

and the regularization parameter was set to $\varepsilon = 0.0$. The final fused output was obtained as $q_i = \bar{a}_i I_i + \bar{b}_i$, clipped to the range [0, 255] and saved as a single-channel grayscale image. This process preserved the thermal saliency from $p$ while enhancing structural detail guided by $I$. Computationally, the RGIF method operates efficiently, with the ECC alignment running at a complexity of $O(N \cdot iters)$, while the guided filtering step is $O(N)$ due to its box-filter implementation. Although CNN-based guided fusion models like GuidedNet [43] perform well, they rely on pixel-level alignment, paired training data, and higher computational cost. In contrast, RGIF is training-free, runs in linear time $O(N)$, and offers low-latency, high-FPS operation robust to cross-modal misalignment, making it suitable for real-time UAV detection.

**Figure 3** presents the workflow of the proposed Registration-aware Guided Image Fusion (RGIF). **Figure 4** illustrates the RGIF-fused image obtained by integrating complementary information from the input modalities.

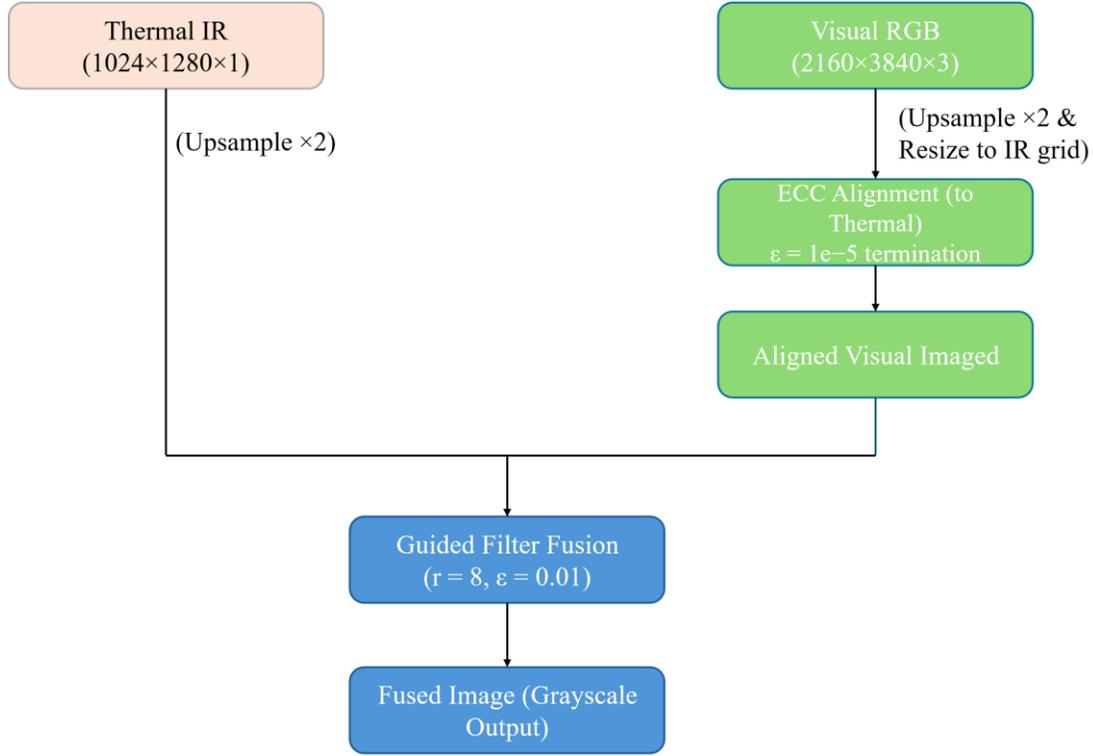

Figure 3: Workflow of the proposed Registration-aware Guided Image Fusion (RGIF)

### 3.3.2. Reliability-Gated Modality-Attention Fusion (RGMAF)

We introduce a two-branch fusion strategy, termed Reliability-Gated Modality-Attention Fusion (RGMAF), to combine thermal and visual images for robust object detection. The process begins with geometric alignment, where the visual frame is warped into the thermal coordinate space using an affine registration based on the enhanced correlation coefficient (ECC) [44]. Alternatively, ORB with RANSAC-based homography was employed [45], and an optional dense optical flow refinement can be enabled to further improve local correspondence. This produced a warped visual image $V_w$ that matched the resolution and perspective of the thermal input I.

Aligned thermal and visual images were then passed through two independent YOLO-style backbones, yielding multi-scale feature maps $F_{\text{thermal}}$ and $F_{visual}$. From these features, per-pixel energy maps were computed and normalized. A SoftMax-based attention mechanism [46] assigned pixel-wise weights as

$$[w_{\text{thermal}}, w_{\text{visual}}] = \text{softmax}(\frac{E_{\text{thermal}}+\beta}{T}, \frac{E_{visual}}{T}) \qquad (4)$$

where β adjusted the weighting toward thermal features and T controlled the sharpness of the distribution. Thermal cues, which provide strong target contrast under diverse illumination conditions, were therefore preserved, while visual features contributed complementary structural information that enhanced localization and scene understanding [41].

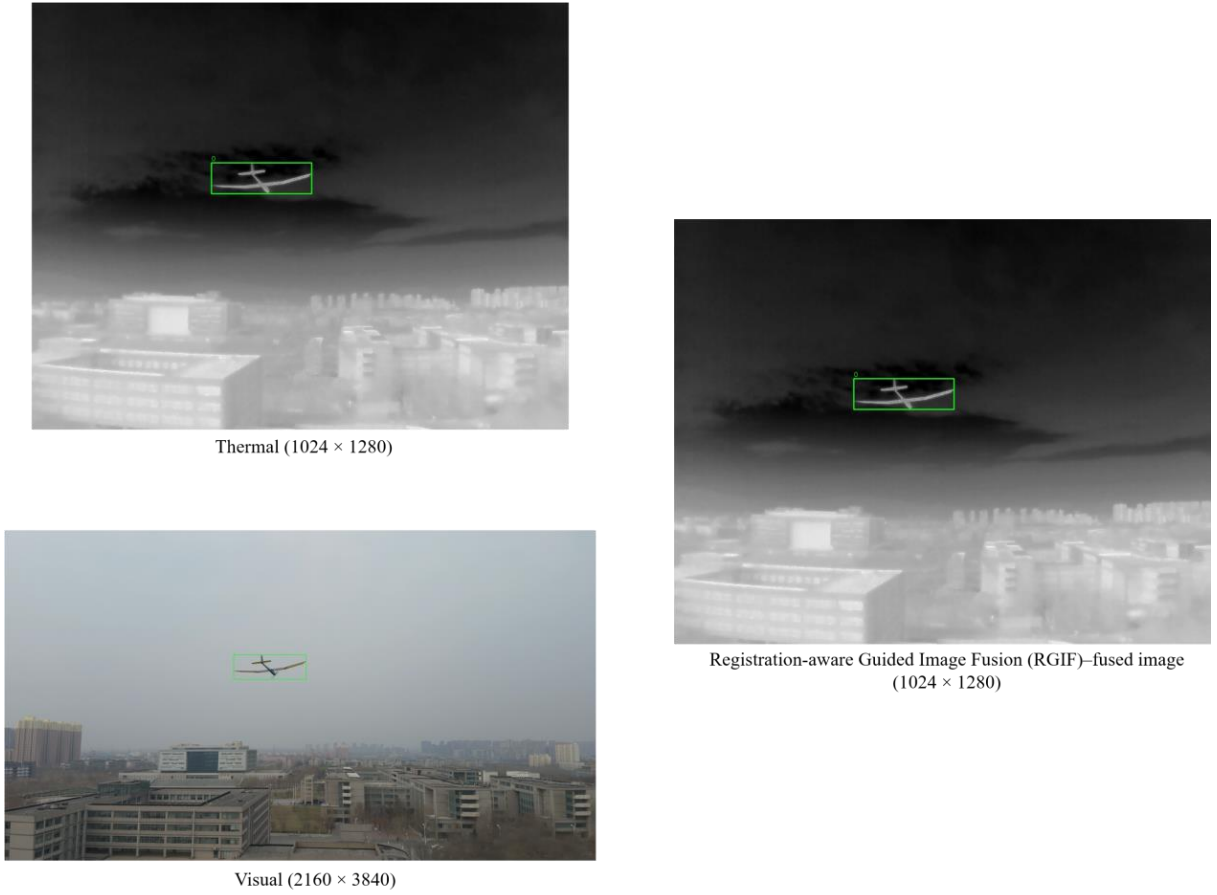

Figure 4: Example of input images and the corresponding RGIF-fused result.

To suppress mis-registration and ghosting, the visual attention map was further modulated by a reliability gate computed from local normalized cross-correlation (NCC) and edge-direction consistency:

$$r = smooth\ (NCC\ (I, V_w)^\gamma \cdot \text{EdgeCons}(I, V_w)),\ W^*_{visual} = W_{visual} \cdot r \tag{5}$$

This ensures that visual information contributes only regions with strong local correspondence.
Fusion was performed in the luminance domain using a base–detail decomposition [47]. Both modalities were smoothed with a Gaussian kernel to obtain base layers, and the residuals defined their respective detail components. The fused luminance was constructed as

$$Y_F = B_I + (D_I + w_{visual} \cdot D_V^{\text{clip}}) \tag{6}$$

where $D_I$ and $D_V$ were the detail layers of the thermal and visual images, respectively, and $D_V^{\text{clip}}$ was clipped to suppress noise. A non-darkening constraint was applied so that $Y_F \geq I$, ensuring that thermal saliency was always maintained in the fused image. A narrow warp-edge region of the visual frame was also suppressed to avoid artifacts introduced by geometric warping.

The final fused output was retained in grayscale form (three identical channels) for compatibility with the detection framework. By preserving thermal contrast as the foundation and integrating visual structures that enrich spatial detail, the method produced fused images that combined the strengths of both modalities—thermal robustness in challenging environments and visual clarity for precise spatial definition—resulting in outputs that were clear, informative, and highly effective for object detection. **Figure 5** presents the overall pipeline of the proposed Reliability-Gated Modality-Attention Fusion (RGMAF) framework. **Figure 6** shows the fused image generated using the Reliability-Gated Modality-Attention Fusion (RGMAF) method, where modality-specific features are adaptively weighted to produce unified representation.

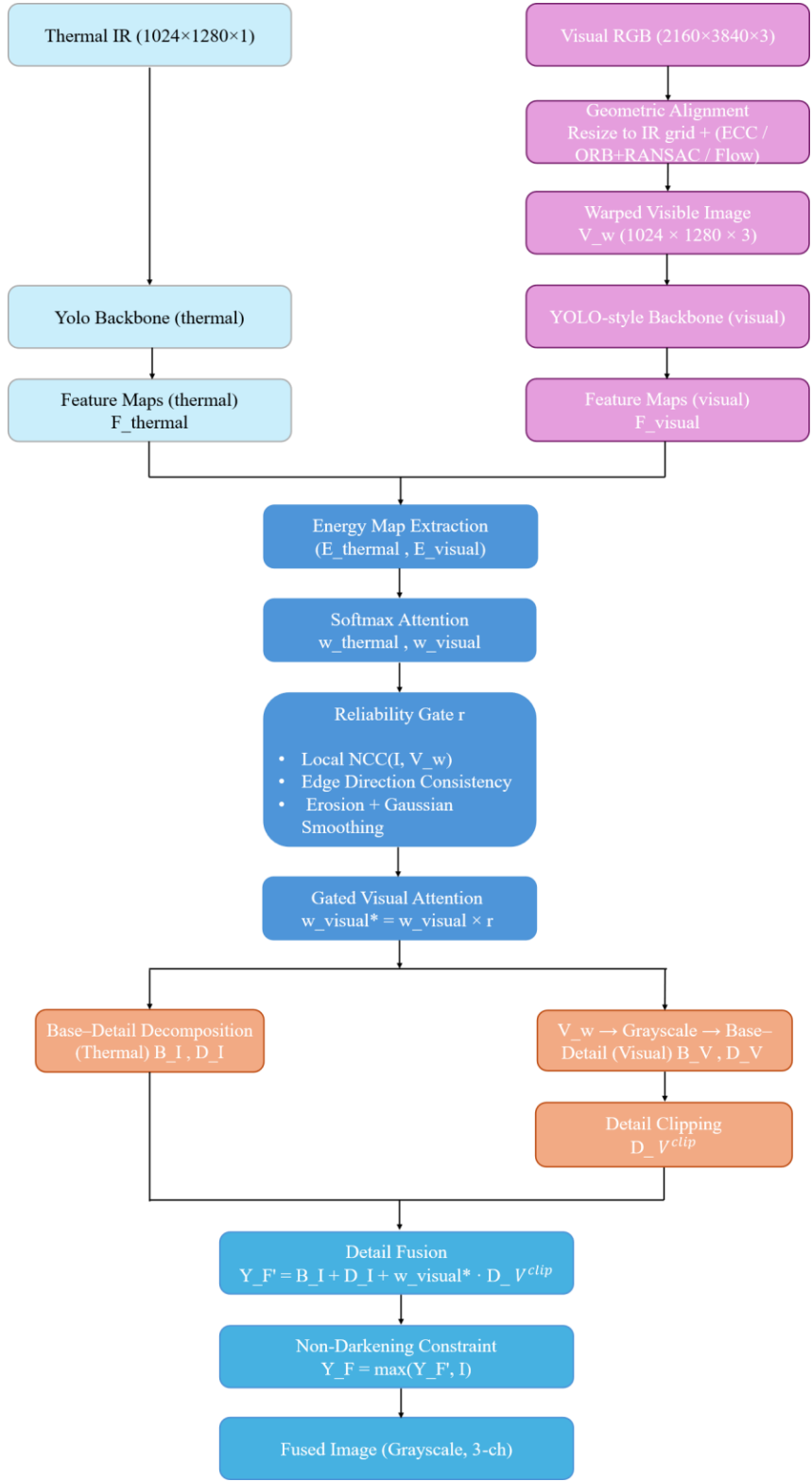

Figure 5: Workflow of the proposed Reliability-Gated Modality-Attention Fusion (RGMAF) framework

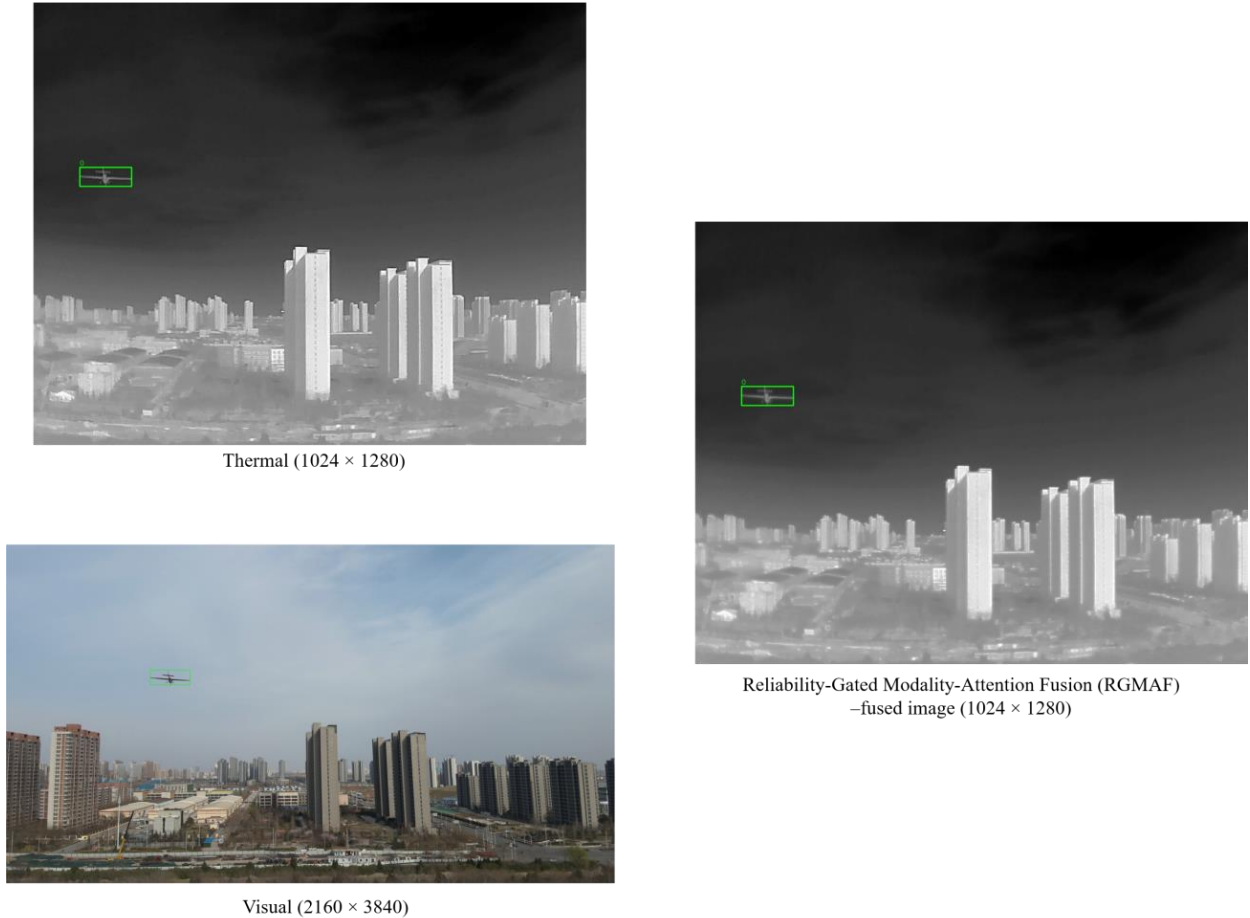
Figure 6: Example of input images and the corresponding RGMAF-fused result.

**3.4. Training and Testing Methodology**
The infrared (thermal) and wide (visual) image datasets were partitioned using group five-fold cross-validation. In this setup, the data corresponding to a single fixed-wing UAV was assigned exclusively to either the training, validation, or test set [48]. This grouping strategy was used to avoid UAV-identity leakage, ensuring that no images from the same aircraft appeared in multiple splits and preventing artificially inflated performance. For each fold, 80% of the data was allocated to the training set and 20% to the test set. From the training portion, an additional 10% was set aside for validation. The reported performance represents the mean across all five folds. The training, validation, and test splits were fixed across all experiments and modalities to ensure consistent and fair comparison. In separate experiments, we evaluated multiple YOLO (You Only Look Once) variants [49]. Initially, several models were examined on the infrared dataset, among which YOLOv9e, YOLOv10x, and YOLOv12x achieved the best performance. These three top-performing models were then further evaluated on the wide (visual) dataset to verify consistency across modalities. Across both datasets, YOLOv10x consistently outperformed the other variants, which motivated its exclusive use in the fusion dataset experiments. All models were initialized using the default COCO-pretrained YOLO weights provided by Ultralytics (pretrained=True), and no custom pretraining strategy was applied. For the Wide (fine-tuned) experiment, the detection model was first trained on the wide-view visual images. The resulting weights were then used to initialize a second training stage on the zoom-view samples using the same training configuration, including the number of epochs and learning rate. The fine-tuned model was subsequently evaluated on the wide-view visual test set.

For all experiments, training was conducted with a predetermined maximum of 200 epochs, an early-stopping patience of 15 epochs, and a batch size of 8. The Stochastic Gradient Descent (SGD) optimizer was employed with learning rate of 0.01, momentum 0.937, and weight decay 0.0005, aligning with widely adopted YOLO optimization settings. Data augmentation was handled through the Ultralytics YOLO augmentation pipeline, which includes mosaic augmentation (disabled in the final training phase), random horizontal flips, HSV-based color jittering, and random affine transformations. In the wide (visual) dataset experiments, we further examined the effect of the rectangular training mode (rect=True vs. rect=False).

The experiments were executed on a workstation equipped with an NVIDIA GeForce RTX 4090 GPU and an AMD Radeon™ GPU, along with 128 GB of RAM and an AMD Ryzen 9 7950X3D 16-core processor. To maintain consistency and reproducibility throughout all folds and model variants, training was performed sequentially on the RTX 4090. All procedures were executed using Visual Studio Code together with the Windows command-line interface. All experiments were conducted using Ultralytics v8.3.83, PyTorch 2.5.0+cu118, and Python 3.11.7.

### 3.5. Evaluation Matrices

To rigorously assess the performance of the detection models, four widely recognized evaluation metrics were employed:

**(a) Precision**: Precision measures the reliability of the model's positive predictions. It is expressed as the proportion of correctly detected objects (True Positives, TP) to the total number of objects predicted as positive (TP + False Positives, FP).

$$P = \frac{TP}{TP+FP} \quad (7)$$

Here, True Positives (TP) refer to objects that are correctly identified by the model, while False Positives (FP) denote instances where the model predicts an object that does not exist in the ground truth.

**(b) Recall**: Recall assesses the model's capacity to detect all relevant objects. It is defined as the ratio of correctly identified objects (True Positives, TP) to the total number of ground-truth objects (TP + False Negatives, FN).

$$R = \frac{TP}{TP+FN} \quad (8)$$

In this context, False Negatives (FN) represent instances where an object is present in the ground truth but was missed by the model.

**(c) Mean Average Precision at IoU = 0.50 (mAP@50)**: This score captures the detection accuracy at a relatively permissive overlap threshold, where predicted and ground-truth bounding boxes are considered correct if they overlap by at least 50%. Formally, if $AP_i^{0.50}$ is the average precision for class i, then:

$$\text{mAP@50} = \frac{1}{N}\sum_{i=1}^{N} AP_i^{0.50} \quad (9)$$

**(d) Mean Average Precision at IoU = 0.50–0.95 (mAP@50–95):** This stricter measure requires the model to perform well under multiple overlap thresholds (from 0.50 to 0.95, in steps of 0.05). It therefore reflects both detection accuracy and precise localization. If $AP_i^t$ denotes the average precision of class i at threshold t, then:

$$\text{mAP@50} - 95 = \frac{1}{N}\sum_{i=1}^{N}\left(\frac{1}{10}\sum_{t=0.50}^{0.95} AP_i^t\right) \quad (10)$$

In addition to the accuracy-based measures, the evaluation also incorporates statistical variability and computational efficiency. All performance metrics are reported as the mean ± standard deviation, computed across the cross-validation folds to quantify both the central tendency and the stability of each model.

**(e) Inference Latency:** Inference latency represents the end-to-end time required to process a single image. It accounts for all stages of the pipeline, including preprocessing, model forward propagation, and post-processing. The total latency is computed as

$$Latency = t_{pre} + t_{inf} + t_{post} \quad (11)$$

where $t_{pre}$, $t_{inf}$, and $t_{post}$ denote preprocessing, inference, and post-processing times, respectively.

**(f) Frames Per Second (FPS):** To quantify real-time performance, the throughput of each model is measured in frames per second and is defined as the inverse of latency:

$$FPS = \frac{1000}{Latency} \quad (12)$$

with latency expressed in milliseconds per image. Higher FPS values indicate stronger suitability for real-time deployment.

These computational efficiency measures complement the accuracy metrics by providing a holistic evaluation of each model's practicality in UAV-based operations, where real-time processing is often required.

## 4. Results

**Table 3** summarizes the quantitative performance of the evaluated architectures across the infrared, wide, and fused modalities. All results are reported as mean ± standard deviation across five cross-validation folds, providing a statistically reliable assessment of each model's stability and generalization capability.

Table 3: Evaluation of baseline models and proposed multimodal fusion techniques, reported with mean ± standard deviation of detection metrics and inference speed (latency and FPS)

| Experiment | Architecture | Precision (±std) | Recall (±std) | mAP50 (±std) | mAP50–95 (±std) | Latency (ms/img) | FPS |
|---|---|---|---|---|---|---|---|
| Infrared | Yolov9e [50] | 98.07 ± 0.47 | 98.03 ± 0.53 | 99.33 ± 0.27 | 87.17 ± 2.69 | 3.30 | 302.6 |
| | Yolov12x [51] | 98.77 ± 0.47 | 98.29 ± 0.96 | 99.00 ± 0.39 | 87.93 ± 2.36 | 3.45 | 290.0 |
| | Yolov10x [52] | 99.19 ± 0.38 | 98.58 ± 1.34 | 99.17 ± 0.42 | 88.48 ± 2.56 | 2.08 | 480.3 |
| Wide | Yolov9e [50] | 91.20 ± 4.08 | 90.89 ± 3.78 | 94.05 ± 3.96 | 72.00 ± 7.87 | 4.80 | 208.0 |
| | Yolov12x [51] | 92.16 ± 4.95 | 88.63 ± 6.35 | 93.41 ± 3.62 | 69.35 ± 5.64 | 4.34 | 230.2 |
| | Yolov10x [52] | 94.06 ± 2.62 | 91.91 ± 5.17 | 95.45 ± 2.54 | 76.59 ± 3.62 | 2.25 | 444.4 |
| Wide (fine-tuned) | Yolov10x [52] | 96.10 ± 3.98 | 94.13 ± 5.06 | 96.65 ± 2.67 | 86.82 ± 4.63 | 2.10 | 476.6 |
| **Proposed (Registration-aware Guided technique)** | Yolov10x [52] | 97.75 ± 1.93 | 94.82 ± 1.22 | 97.65 ± 0.67 | 84.96 ± 4.13 | 2.07 | 482.0 |
| **Proposed (Reliability-Gated Modality-Attention)** | Yolov10x [52] | 98.64 ± 0.80 | **98.64 ± 1.15** | 99.10 ± 0.55 | 88.01 ± 3.07 | 3.10 | 322.0 |

### 4.1 Infrared Modality Performance

Across the infrared modality, YOLOv9e, YOLOv10x, and YOLOv12x demonstrated superior detection accuracy relative to the remaining architectures. YOLOv10x exhibited the most balanced and consistent performance, achieving 99.19 ± 0.38% precision, 98.58 ± 1.34% recall, 99.17 ± 0.42% mAP@50, and 88.48 ± 2.56% mAP@50–95. These results indicate strong localization reliability, even under stricter IoU thresholds.

Although YOLOv9e attained the highest mAP@50 (99.33 ± 0.27%), its reduced mAP@50–95 (87.17 ± 2.69%) and higher variability suggest comparatively weaker bounding-box stability. YOLOv12x showed similar trends—high mAP@50 (99.00 ± 0.39%) paired with lower mAP@50–95 (87.93 ± 2.36%)—reflecting limited robustness at fine-grained localization.

From an efficient standpoint, YOLOv10x achieved the lowest latency (2.08 ms per image) and the highest throughput (480.3 FPS) among the infrared models, confirming that its superior accuracy is coupled with outstanding real-time capability. These results establish YOLOv10x as the most reliable infrared baseline in both performance and computational efficiency.

### 4.2 Wide Modality Performance

Detection accuracy decreased substantially when evaluated on the wide-field modality. The increased field of view and reduced target size resulted in higher prediction variability. YOLOv9e and YOLOv12x recorded modest performance, with mAP@50–95 values of 72.00 ± 7.87% and 69.35 ± 5.64%, respectively, indicating relatively weak localization under wide imaging conditions.

YOLOv10x performed comparatively better, achieving 94.06 ± 2.62% precision, 91.91 ± 5.17% recall, and 76.59 ± 3.62% mAP@50–95, while maintaining a latency of 2.25 ms (444.4 FPS). This suggests improved robustness but still highlights the inherent difficulty of the wide modality.

When fine-tuned with zoom-informed wide-field data, YOLOv10x showed substantial gains, achieving 96.10 ± 3.98% precision, 94.13 ± 5.06% recall, 96.65 ± 2.67% mAP@50, and 86.82 ± 4.63% mAP@50–95, while sustaining a low latency of 2.10 ms (476.6 FPS). These improvements demonstrate the effectiveness of modality-specific adaptation and highlight the benefits of incorporating wide-zoom training samples.

**4.3 Fusion-Based Performance: RGIF and RGMAF**

The proposed fusion frameworks further improved performance beyond the wide-only and fine-tuned baselines. The Registration-aware Guided Image Fusion (RGIF) method leveraged geometric alignment between modalities to obtain 97.75 ± 1.93% precision, 94.82 ± 1.22% recall, 97.65 ± 0.67% mAP@50, and 84.96 ± 4.13% mAP@50–95. Notably, RGIF also achieved the lowest inference latency and the highest processing speed among all evaluated configurations—including the infrared, wide, fine-tuned, and fused models—operating at 2.07 ms per image and 482 FPS. This highlights that precise geometric alignment preserves exceptional computational efficiency, making RGIF highly suitable for real-time UAV deployment.

The Reliability-Gated Modality-Attention Fusion (RGMAF) achieved the best overall performance, with 98.64 ± 0.80% precision, 98.64 ± 1.15% recall, 99.10 ± 0.55% mAP@50, and 88.01 ± 3.07% mAP@50–95. Although RGMAF incurs a higher computational cost (3.10 ms latency, 322 FPS), it provides the highest recall and the most consistent detection stability across metrics. The slightly higher latency of RGMAF can be attributed to the inherent complexity of its fusion mechanism. In contrast to simpler concatenation-based or single-stream detectors, RGMAF performs reliability estimation, gated feature weighting, and attention-based modality integration. These additional steps introduce extra matrix operations and dynamic computations during inference, resulting in a moderate increase in processing time. However, this computational overhead is expected given the richer cross-modal reasoning the model performs.

**Qualitative Analysis of Single-Modality Degradation**

We further perform a robust analysis using the Reliability-Gated Modality-Attention Fusion (RGMAF) strategy, as it represents the best-performing fusion configuration in our experiments. To evaluate robustness under modality degradation, we conducted controlled experiments in which one sensing modality was intentionally degraded prior to fusion, while the other modality was kept unchanged. The degraded modality was either wide-view visual or thermal infrared. Gaussian blurring with a kernel size of 15×15 was applied to both images, followed by intensity scaling with a factor of 0.6 for the visual image to simulate reduced image sharpness and contrast. **Figure 7** further provides a qualitative illustration of image degradation across both modalities.

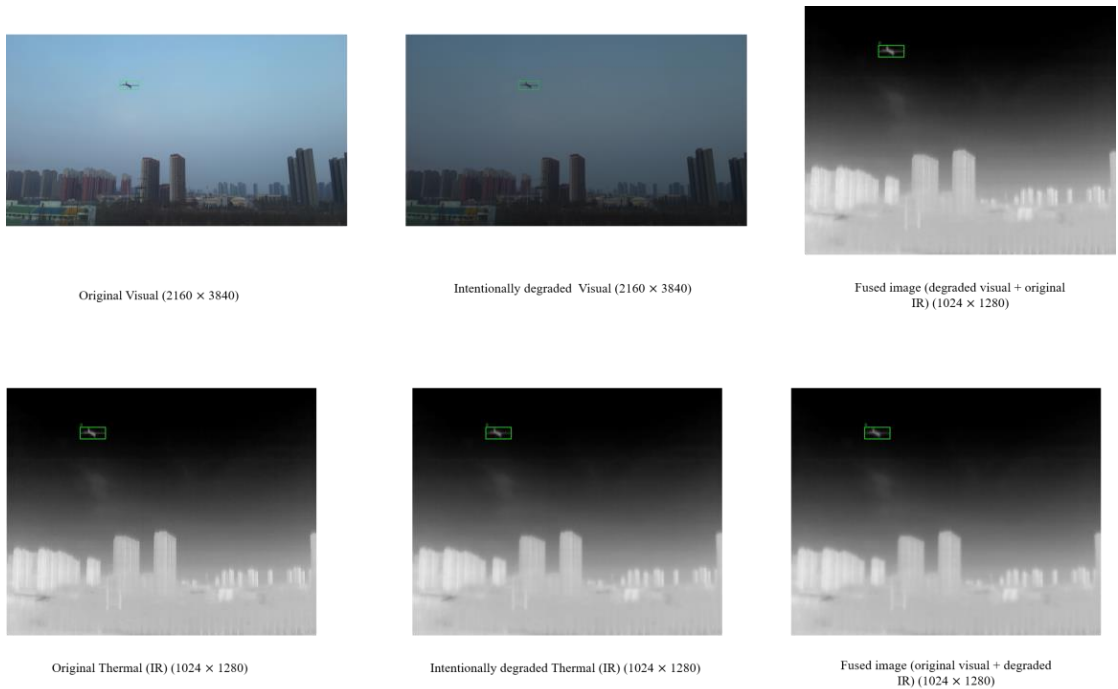

Figure 7: Qualitative example of visual–infrared fusion under visual degradation

Degradation was applied only during evaluation and did not affect training. All robustness experiments were conducted using the previously trained YOLOv10x model and the same test set. As shown in **Table 4**, degrading either modality results in a limited performance change, and detection performance does not collapse in either case. This behavior indicates graceful degradation and demonstrates that the proposed fusion strategy remains effective under realistic sensor degradation scenarios.

Table 4: Robustness analysis under controlled modality degradation

| Condition | Precision (%) | Recall (%) | mAP@50 (%) | mAP@50–95 (%) |
|---|---|---|---|---|
| **Normal RGMAF fusion** | 98.63 | 98.64 | 99.10 | 88.02 |
| **Visual-degraded RGMAF fusion** | 98.60 | 98.03 | 98.90 | 87.02 |
| **Infrared-degraded RGMAF fusion** | 98.06 | 97.47 | 98.54 | 82.71 |

## 4.4 Convergence Curves

**Figure 8** - **Figure 9** present the training and validation curves of the infrared and wide-modality detectors. The infrared model illustrates stable convergence across all monitored metrics, with losses decreasing consistently and precision and recall rapidly surpassing 0.99. The corresponding mAP@50 stabilized at approximately 99.5%, while mAP@50–95 approached 91%, indicating effective learning and strong generalization without signs of overfitting. The wide-modality detector also demonstrates consistent learning and generalization, as reflected by steadily decreasing losses and improving precision and recall. However, its mAP@50–95 remained lower than that of the infrared model, reflecting weaker bounding-box localization.

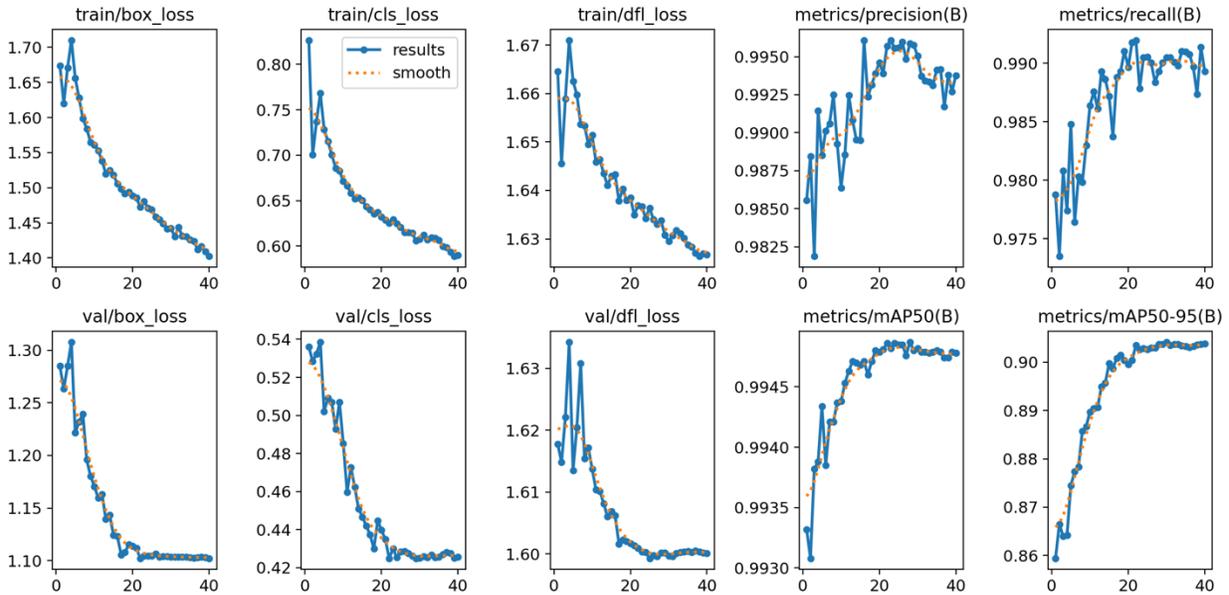

Figure 8: Training and validation curves for thermal (infrared) images with YOLOv10x

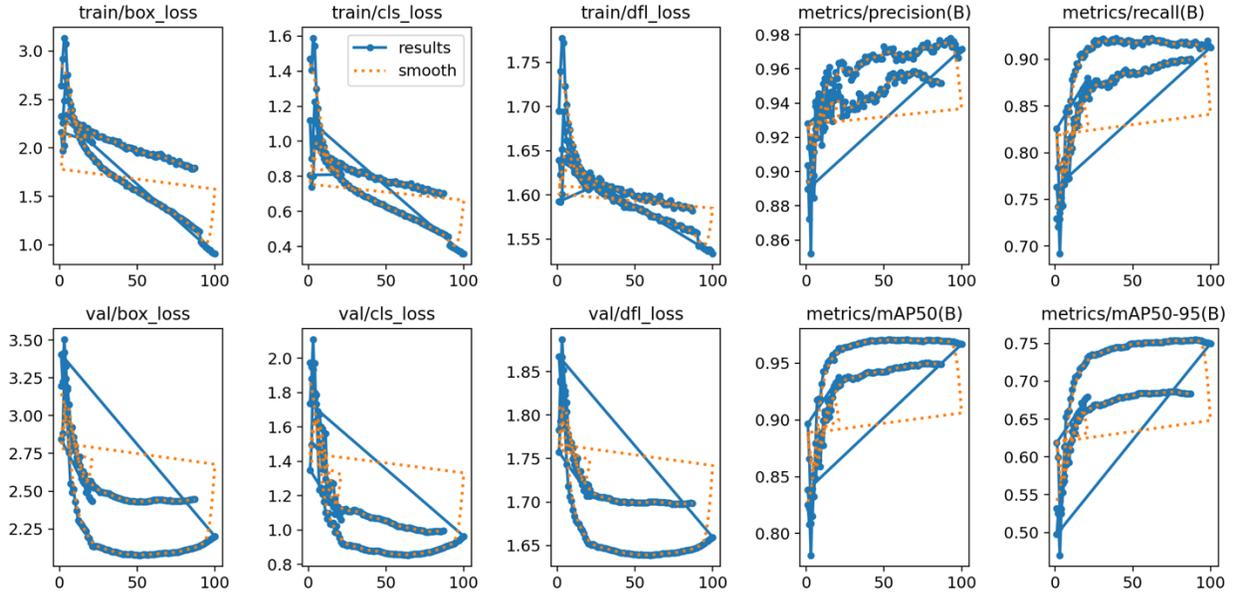

Figure 9: Training and validation curves for visual (wide) modality using YOLOv10x

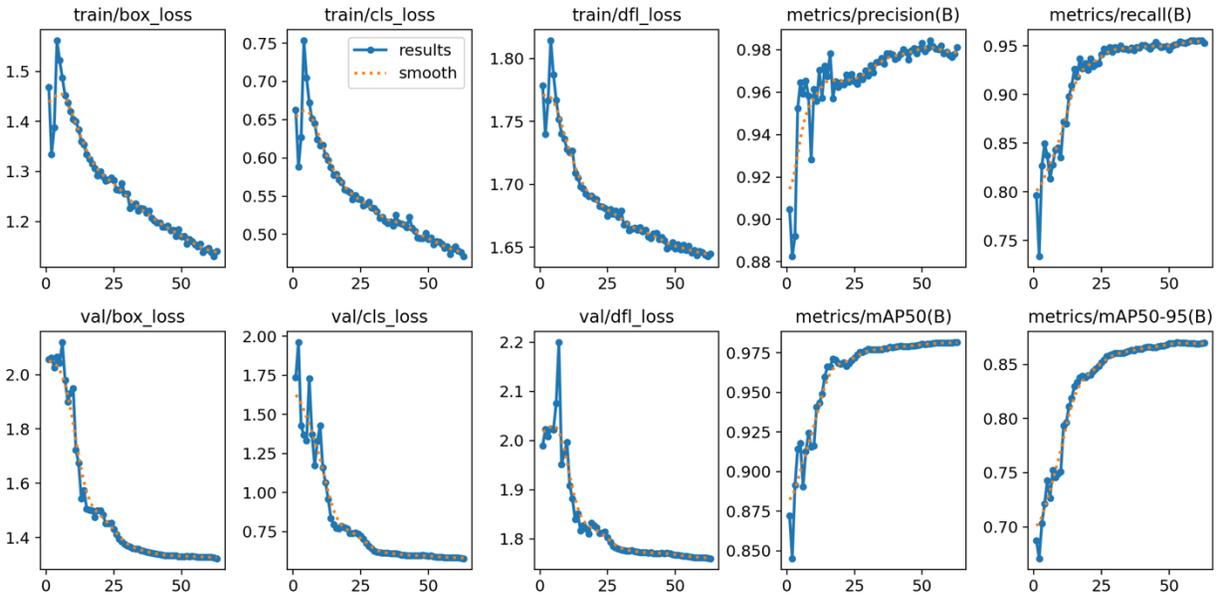

Figure 10: Training and validation curves for the wide model fine-tuned on zoom images using YOLOv10x

**Figure 10** shows the training behavior of the wide-tuned detector, which achieved more stable convergence than the baseline wide model. All loss components decreased consistently, while precision and recall improved rapidly and stabilized above 0.95. The mAP@50 approached 0.98 and mAP@50–95 converged to approximately 0.87, confirming that fine-tuning with wide-zoom data significantly enhanced generalization and reduced the performance gap relative to the infrared modality.

**Figure 11** depicts the curves of the proposed Registration-aware Guided Image Fusion (RGIF) model. The model converged smoothly, with decreasing losses and continuous improvements in precision, recall, and mAP values. Compared with the wide-only and wide-tuned detectors, R-GIF achieved higher recall and mAP@50–95, demonstrating improved localization robustness. Although the infrared model retained a slight advantage in mAP@50–95, the difference was marginal and did not outweigh the advantages of fusion.

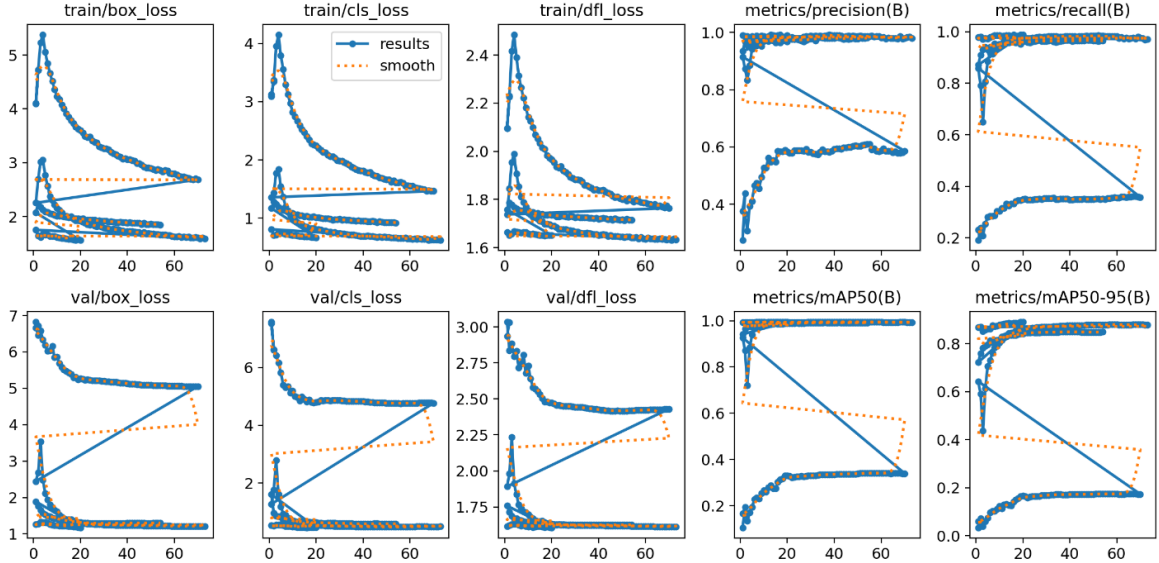

Figure 11: Training and validation curves for the Registration-aware Guided Image fusion technique using YOLOv10x

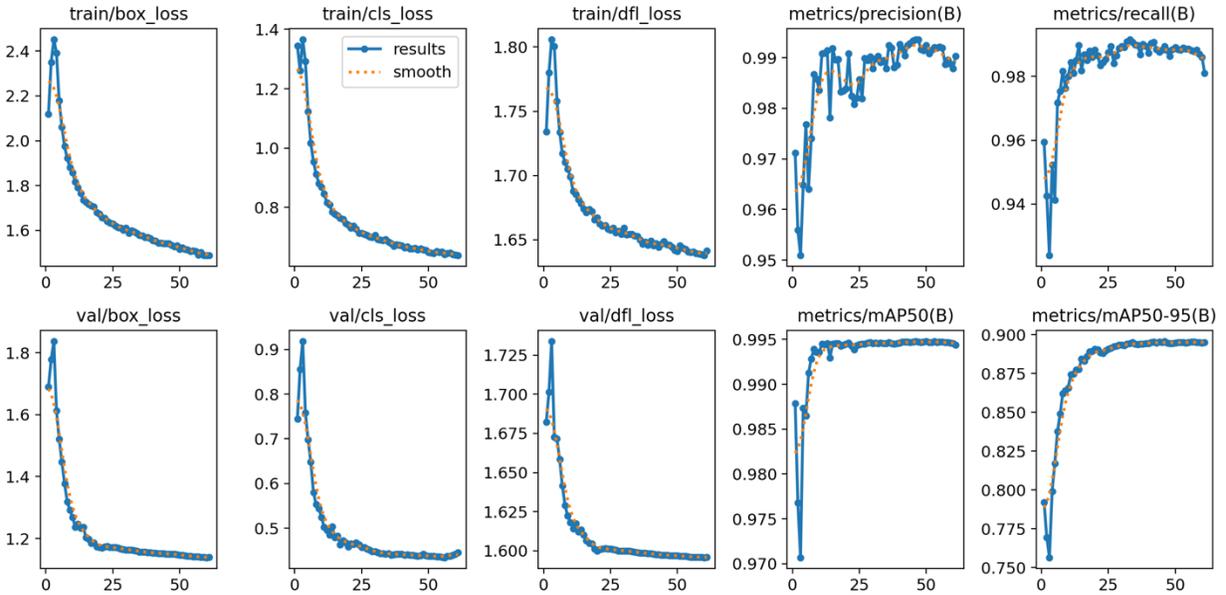

Figure 12: Training and validation curves for the Reliability-Gated Modality-Attention fusion technique using YOLOv10x

Finally, **Figure 12** presents the Reliability-Gated Modality-Attention (RGMAF) model, which converged rapidly and reached very high levels of precision, recall, and mAP. Both mAP@50 and mAP@50–95 plateaued at strong values, confirming that RGMAF achieved the best balance between detection accuracy and recall. In comparison with the wide-only, wide-tuned, and infrared models, RGMAF demonstrated superior robustness and consistency, establishing it as the most effective approach among all evaluated strategies.

### 4.5 Ablation Study
To further investigate the impact of different architectural choices and fusion strategies, we conducted a set of ablation experiments. The results are summarized in **Table 5**. For the baseline detectors, earlier YOLO variants such as YOLOv5m and YOLOv5l achieved relatively poor performance, with mAP@50–95 values of 54.52% and 53.53%, respectively. In contrast, more recent architectures delivered substantial improvements on the infrared modality. In particular, YOLOv10l achieved 99.08% mAP@50 and 87.62% mAP@50–95, while YOLOv11x and YOLOv12l maintained similar levels of accuracy, indicating that architectural advances beyond YOLOv8 lead to consistently stronger detection performance. Compared with these models, YOLOv9e, YOLOv10x, and YOLOv12x obtained better results across the evaluation metrics, and were therefore identified as the three most effective models. Based on this observation, we selected these three architectures for subsequent experiments, as training multiple models is computationally demanding and time-consuming. For the wide modality, YOLOv10x with reactive training enabled (rect=True) achieved 93.28% precision, 93.08% recall, and 95.59% mAP@50, while its mAP@50–95 remained limited at 75.08%, confirming the inherent difficulty of this modality compared to infrared. The rect=True option minimizes padding by maintaining the original aspect ratio of the wide-view images during training, which is particularly relevant given their elongated spatial structure. This adjustment led to a slight improvement in mAP@50 compared with rect=False; however, the stricter mAP@50–95 metric was approximately 1% lower. Since mAP@50–95 provides a more reliable indication of localization accuracy, the configuration with rect=False was ultimately preferred in later experiments. All ablation models were trained using identical augmentation settings, batch size, and training schedule to ensure a fair comparison. Only the architectural or fusion components were varied.

In Section 0, we outlined seven traditional image fusion techniques commonly applied to same-sized images. We implemented these approaches—alpha blending, weighted averaging, overlay fusion, Laplacian pyramid, wavelet transform fusion, guided filtering, and YCrCb fusion—on our dataset. However, as our infrared and visual images differ in size and resolution, these methods were not directly suited to our setting, and their limitations became evident during implementation. As shown in **Figure 13**, only the wavelet and guided-based methods produced fused images in which object alignment appeared reasonably, whereas the remaining techniques failed to merge structural details effectively.

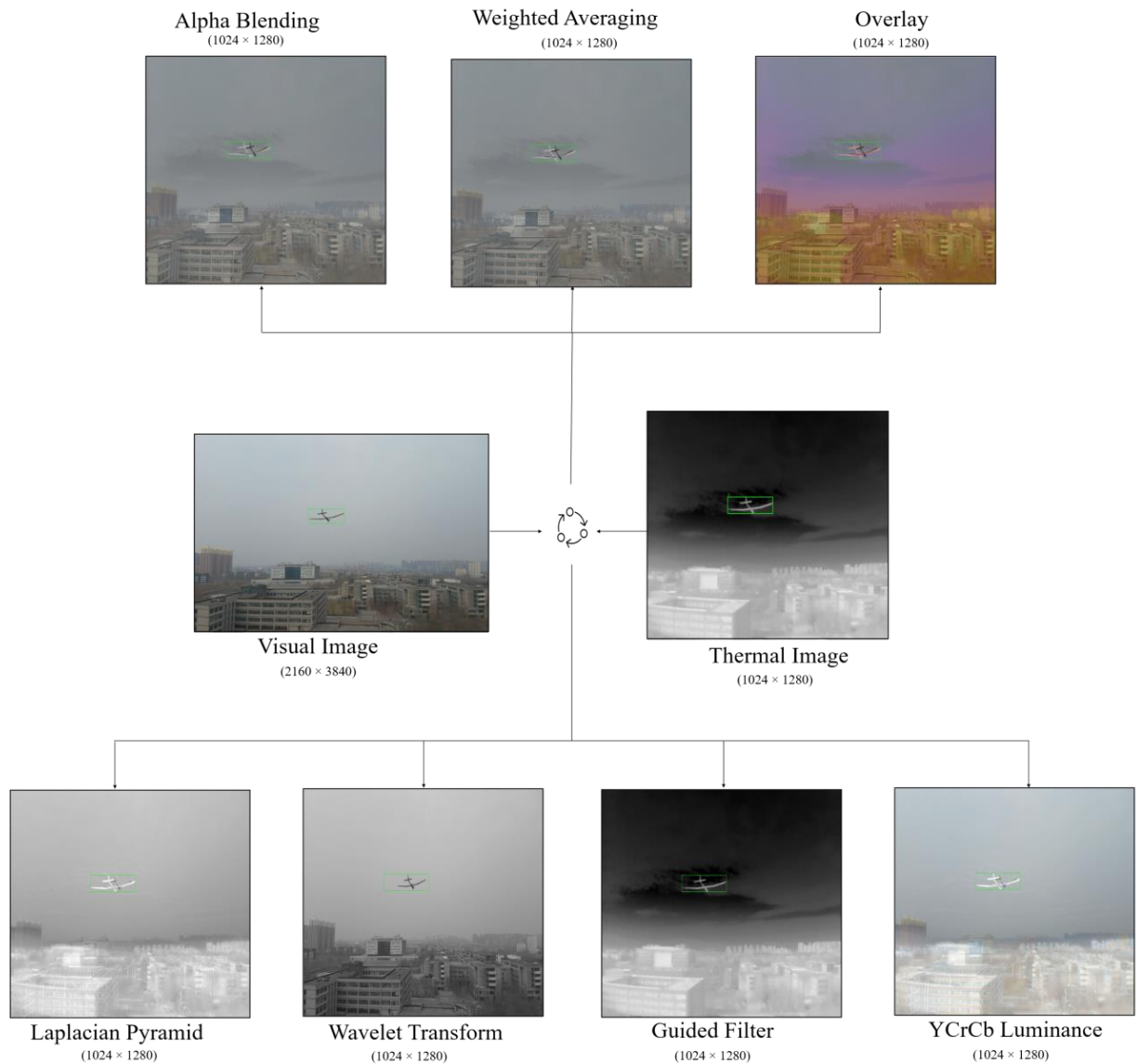

Figure 13: Visual results of traditional image fusion approaches

For this reason, the wavelet method was selected as a representative baseline for further evaluation. Quantitative results showed that wavelet fusion achieved 80.19% recall and 70.78% mAP@50–95, which is considerably lower than the infrared baseline. Moreover, inspection at very high resolution revealed subtle artifacts in the wavelet-fused images, such as faint shadows reflecting misalignment of object boundaries, which were not visible at normal scales. A small-scale manual review showed that one out of four inspected wavelet-fused samples (≈25%) exhibited visible ghosting or mild boundary drift. Although this rate is modest, these artifacts still produce structural inconsistencies that negatively impacted downstream detection. We also investigated the guided filter; however, the generated images showed misalignment between objects and their bounding box annotations. guided-filter fusion exhibited an estimated ≈ 43% misalignment rate, including structurally severe cases where object boundaries appeared noticeably shifted relative to the ground-truth labels. Even when the fused images appeared visually plausible, the geometric inconsistency made them unsuitable for reliable object detection. As a result, the guided filter, along with other classical fusion methods, was not pursued in subsequent experiments. As illustrated in **Figure 14**, wavelet fusion introduced subtle artifacts, while guided filtering led to noticeable misalignment issues.

Finally, a decision-based fusion experiment was performed, in which predictions from infrared and wide models were combined at the decision level. Although this approach increased recall (80.33%), precision dropped significantly to 61.60%, and the resulting mAP@50–95 was only 69.72%. These results highlight the limitations of naive late fusion, where error accumulation and inconsistencies between modalities degrade overall performance. In particular, because the infrared and wide images differ in size and resolution, the decision process was biased toward the stronger infrared predictions, while the weaker wide predictions reduced overall accuracy. This imbalance reflects both the resolution mismatch between modalities and the shortcomings of the decision-level strategy.

Collectively, the ablation results underscore two key points:
- Modern YOLO architecture demonstrated substantial performance gains over earlier versions, making it unnecessary and not practically wise to extend wide-modality experiments with older models.
- Traditional fusion approaches are not well suited for images of differing sizes, with both wavelet-based and decision-level fusion proving insufficient to achieve competitive results.

These findings highlight the need for more principled strategies, motivating the development of the proposed Registration-aware Guided Image Fusion (RGIF) and Reliability-Gated Modality-Attention Fusion (RGMAF), which are specifically designed to address these limitations and enable robust multimodal detection in the context of heterogeneous sensor image fusion.

Table 5: Performance impact of individual components in the ablation study

| Experiment | Architecture | Precision (%) | Recall (%) | mAP@50 (%) | mAP@50–95 (%) |
|---|---|---|---|---|---|
| Infrared | Yolov5m [53] | 87.68 | 64.05 | 77.21 | 54.52 |
| | Yolov5l [53] | 89.84 | 60.66 | 75.79 | 53.53 |
| | Yolov8m [54] | 97.83 | 96.88 | 98.36 | 86.03 |
| | Yolov8l [54] | 98.38 | 97.12 | 98.54 | 86.64 |
| | Yolov8x [54] | 97.28 | 96.03 | 97.82 | 83.44 |
| | Yolov10l [52] | 98.87 | 98.38 | 99.08 | 87.62 |
| | Yolov11x [55] | 99.03 | 98.42 | 98.17 | 87.99 |
| | Yolov12l [51] | 98.74 | 98.06 | 98.94 | 87.35 |
| Wide | Yolov10x [52] (react=true) | 93.28 | 93.08 | 95.59 | 75.08 |
| Wavelet Fusion | Yolov10x [52] | 88.70 | 80.19 | 84.3 | 70.78 |
| Decision-based Fusion | Yolov10x [52] | 61.60 | 80.33 | 79.0 | 69.72 |

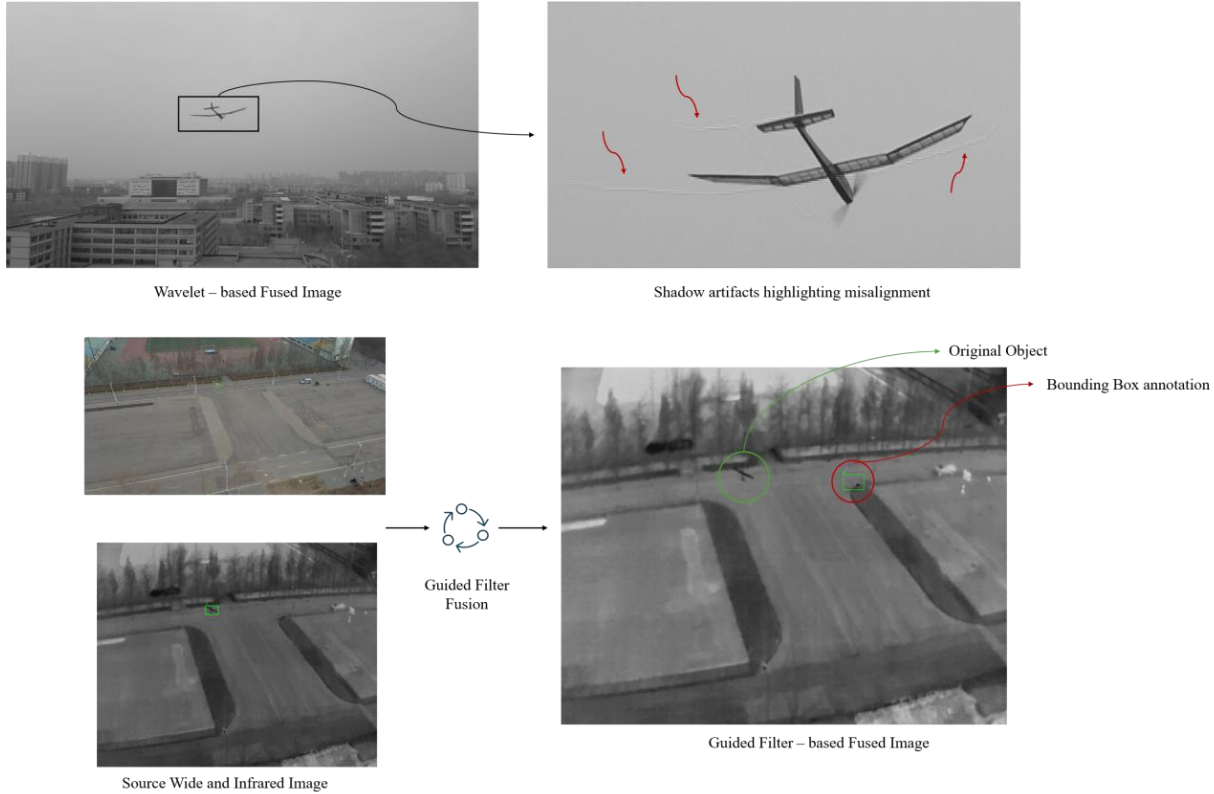

Figure 14: Issues observed with wavelet and guided filter–based fusion methods

## 5. Discussion

The fusion of heterogeneous sensors in UAV detection remains relatively underexplored, especially when the constituent modalities differ substantially in spatial resolution and image dimensions. This limitation is evident across existing IR–VIS fusion studies, where misalignment and scale disparity often constrain downstream performance. In this work, we addressed these challenges through two alignment-oriented fusion strategies: Registration-aware Guided Image Fusion (RGIF) and Reliability-Gated Modality-Attention Fusion (RGMAF). Among the two approaches, RGMAF demonstrates the strongest overall detection performance. This result suggests that reliability-weighted fusion better handles the variability present in UAV sensing environments. In addition to multimodal fusion, we conducted an extensive evaluation of single-modality detectors. Across repeated trials, YOLOv9e, YOLOv10x, and YOLOv12x emerged as the strongest candidates, with YOLOv10x demonstrating the most consistent performance across both infrared and wide-view modalities. Consequently, YOLOv10x was selected as the primary backbone due to its balance between accuracy and computational efficiency. Notably, the infrared detector achieved over 99% mAP@50 while maintaining low inference latency of 2.08 ms and a throughput of 480 FPS. This strong performance is consistent with the larger thermal dataset and the inherently discriminative nature of infrared signatures. In contrast, the visual (wide-view) modality achieved approximately 95% mAP@50 with a latency of 2.25 ms and a throughput of 444 FPS. This reflects the lower separability of visual cues for small UAVs under diverse illumination and background conditions. Wide-view images were selected as the primary visual input because they consistently captured the UAVs in full, whereas zoom-view images frequently contained cropped or partially visible targets. Instead of treating zoom images as an independent modality, they were used to fine-tune the wide-view detector. This approach exploits the higher apparent object resolution available in zoom imagery. This fine-tuning improved performance from 95.45% to 96.65% mAP@50 (a gain of 1.2 percentage points), accompanied by a reduced latency of 2.10 ms and an increased throughput of 476 FPS.

In the fusion experiments, several classical image fusion techniques were first evaluated, including alpha blending, weighted averaging, overlay fusion, Laplacian pyramid fusion, wavelet-transform fusion, guided filtering, and YCrCb-based fusion. None of these approaches produced reliable results on our dataset, primarily due to substantial disparities in image size and spatial resolution between the thermal and visual modalities. In practice, alignment was often achieved for a single UAV while other objects remained misaligned. Even when alignment was successful, the bounding box annotations no longer matched the fused image accurately. These limitations highlighted the need for a dedicated fusion mechanism that enforces global alignment while preserving annotation consistency.

To address this challenge, we introduced Registration-aware Guided Image Fusion (RGIF), which successfully produced spatially coherent fused images while maintaining consistent object alignment. Although its performance did not surpass the thermal modality, RGIF consistently exceeded the visual baseline. It achieved 97.65% mAP@50 and 97.75% precision with an inference latency of 2.07 ms (482 FPS). Building upon this, we further developed Reliability-Gated Modality-Attention Fusion (RGMAF), which adaptively modulates modality contributions based on their reliability. RGMAF achieved 98.64% recall and over 99% mAP@50, with a latency of 3.10 ms (322 FPS). These results surpassed all visual-only and classical fusion approaches, establishing RGMAF as the most effective multimodal integration strategy in our study. Although RGMAF introduces a small increase in latency compared with single-modality detectors, it yields a strong accuracy–efficiency trade-off. By combining thermal and visual cues, the method remains reliable when one modality is weakened by environmental factors. The modest computational overhead is justified by the substantial robustness gains, including a +6.7% improvement in recall over the wide-only baseline and a +0.7% gain over the infrared-only detector.

Beyond numerical performance, the proposed fusion frameworks also address several operational challenges encountered in real-world UAV monitoring scenarios. In practical deployments such as airport perimeter surveillance, border security monitoring, and critical-infrastructure protection, UAVs must be detected under highly variable environmental and sensing conditions. Thermal and visual sensors often experience strong distribution shifts. These include day–night lighting changes, weather-induced visibility degradation, and modality-specific artifacts such as thermal bloom or RGB motion blur. By jointly using complementary cues from both modalities, the proposed RGIF and RGMAF frameworks improve robustness under distribution shifts. If visual imagery becomes unreliable, the thermal modality dominates the fused representation and when thermal contrast weakens visual structures contribute more strongly. This adaptive reliability-based fusion maintains stable detection performance across diverse sensing conditions.

There are relatively few studies that have specifically focused on fusion techniques in the domain of UAV detection. Cheng et al. [25] introduced SLBAF-Net for UAV detection, reporting 0.909 precision and 0.912 recall on a small dual-modal dataset of 2,850 image pairs, though the limited size and quality reduced its reliability. In comparison, our proposed methods were evaluated on a significantly more robust dataset and not only demonstrated superior accuracy and generalizability but also clearly surpassed the results reported by SLBAF-Net. Beyond performance differences, the two approaches also differ fundamentally in their fusion strategies. SLBAF-Net adopts an early, feature-level fusion mechanism that adaptively merges visible and thermal feature maps, implicitly assuming spatially aligned inputs with comparable resolutions. In contrast, our approach is explicitly designed to handle heterogeneous sensing conditions by incorporating alignment awareness and modality reliability gating, enabling robust detection even in the presence of cross-modal misalignment or modality degradation. A comparison table was not included due to the lack of directly comparable studies on heterogeneous thermal-visual fusion for UAV detection. The limited existing works differ substantially in sensing modalities and task objectives, preventing a fair or meaningful tabular comparison.

Although the infrared modality achieved near-saturated performance due to the scale and quality of the dataset, this saturation does not indicate a limitation of the fusion strategies themselves but rather reflects the strength of the baseline. With over 147,000 high-resolution, precisely annotated images, the infrared detector operated close to the upper bound of mAP@50 and mAP@50–95, leaving very limited headroom for further numerical gains. Importantly, the value of the proposed fusion methods lies not in surpassing this saturated baseline but in resolving the challenge of heterogeneous sensor integration and maintaining robustness across varying sensing conditions. Future improvements may incorporate domain adaptation, multimodal contrastive regularization, or feature-space alignment to improve generalization beyond the dataset ceiling.

Looking ahead, several research directions can further strengthen the applicability of the proposed framework. First, evaluating the models on additional UAV datasets with varying sensor characteristics, altitudes, and environmental conditions would help assess generalizability under broader distribution shifts. Second, computational efficiency could be further improved through lightweight reliability-gated modules or parameter-sharing strategies, enabling deployment on resource-constrained UAV hardware. In addition, advanced training optimization methods designed for efficient and stable convergence under imbalanced data, such as the Focal Quotient Gradient System (QGS-Focal) proposed by Lv et al. [56], may offer a complementary direction for further enhancing training robustness and efficiency. Third, extending the framework with temporal reasoning from video streams may enhance robustness to occlusion and motion blur. Finally, transformer-based multimodal encoders, including cross-attention mechanisms and token-level fusion architecture, represent a promising direction. They enable more expressive feature alignment across heterogeneous sensors. Advancing in these directions can enable reliable, real-time multimodal UAV perception suitable for deployment.

## 6. Conclusion

In this study, we demonstrated that heterogeneous infrared–visual fusion can be performed reliably even when the two modalities exhibit substantial differences in resolution and field of view. To address the associated alignment and integration challenges, we introduced two fusion strategies: Registration-aware Guided Image Fusion (RGIF) for pixel-level alignment and Reliability-Gated Modality-Attention Fusion (RGMAF) for adaptive feature weighting. RGIF improved performance over the visual modality, while RGMAF delivered the strongest overall detection performance, yielding a +3.05% gain in mAP@50–95 relative to RGIF and an +17.23% improvement over traditional wavelet-based fusion. Despite incorporating an additional gating mechanism, RGMAF sustained real-time operation at 322 FPS, confirming its suitability for embedded and resource-limited UAV systems. These results demonstrate that carefully designed multimodal fusion remains essential for maintaining detection reliability under diverse illumination, weather, and sensing conditions, and they position the proposed framework as a viable foundation for real-time UAV intrusion monitoring and broader aerial safety applications.

**Declaration:**

**Funding:** The authors declare that no external funding was received for this study.

**Competing Interests:** The authors declare no competing interests.